\documentclass[dvipdft]{article}
\pdfoutput=1

\usepackage{cite}
\usepackage{amsmath,spconf,amssymb,amsfonts}
\usepackage{algorithmic}
\usepackage{textcomp}
\usepackage{xcolor}
\usepackage{soul}
\usepackage{csquotes}
\usepackage{verbatim}

\usepackage{pdfpages}
\usepackage{bbm,dsfont}

\usepackage{comment}
\usepackage{csquotes}
\usepackage{url}
\usepackage{tabularx}

\title{Preservation of Anomalous Subgroups On Machine Learning Transformed Data}
\name{%
\begin{tabular}{@{}c@{}}
Samuel C.~Maina$^{1}$ \qquad 
Reginald E.~Bryant$^{1}$ \qquad 
William O.~Ogallo$^{1}$\\ 
Robert-Florian Samoilescu$^{3}$ \qquad 
Kush R.~Varshney$^{2}$ \qquad 
Komminist Weldemariam$^{1}$
\end{tabular}}
 \address{
 $^{1}$ IBM Research, Nairobi, Kenya \\
 $^{2}$IBM Research, Yorktown Heights, NY, USA\\
 $^{3}$Politehnica University of Bucharest, Bucharest, Romania}

\makeatletter
\def\ScaleIfNeeded{%
\ifdim\Gin@nat@width>\linewidth
\linewidth
\else
\Gin@nat@width
\fi
}
\makeatother
\begin{document}
\ninept
\maketitle

\begin{abstract}
In this paper, we investigate the effect of machine learning based anonymization on anomalous subgroup preservation. In particular, we train a binary classifier to discover the most anomalous subgroup in a dataset by maximizing the bias between the group’s predicted odds ratio from the model and observed odds ratio from the data. We then perform anonymization using a variational autoencoder (VAE) to synthesize an entirely new dataset that would ideally be drawn from the distribution of the original data. We repeat the anomalous subgroup discovery task on the new data and compare it to what was identified pre-anonymization. We evaluated our approach using publicly available datasets from the financial industry. Our evaluation confirmed that the approach was able to produce synthetic datasets that preserved a high level of subgroup differentiation as identified initially in the original dataset. Such a distinction was maintained while having distinctly different records between the synthetic and original dataset.  Finally, we packed the above end-to-end process into what we call Utility Guaranteed Deep Privacy (UGDP) system.  UGDP can be easily extended to onboard alternative generative approaches such as GANs to synthesize tabular data. 
\end{abstract}
\section{Introduction}
\label{sec:intro}

Data-driven processes of identifying interesting and useful subgroups in populations present an ongoing challenge. For example, financial services providers are increasingly relying on machine learning-based approaches for subgroup discovery in shaping investment and marketing strategies \cite{Lopez2019, SunMXZX2014},  designing bespoke portfolio and insurance products \cite{LambertonBH2017,Thesmar2019}, managing risk \cite{Liebergen2017}, detecting fraud \cite{SharmaP2012}, and complying with anti-discrimination or fairness regulations \cite{Bruckner2018}. In healthcare, attempts have been made in identifying and classifying patients' subgroups and clusters with similar prognoses and manifestations as well as responses to different treatment regimes in an attempt at personalized medical service provision \cite{Chen2014,Knevel2019,Ferreira2018,Nielsen2017}. A key objective in these tasks is the discovery of customer or patients' segments, clusters or subgroups in individual-level data, defined by demographic, psychographic, behavioral,  or other variables, that are interesting or anomalous according to some criterion \cite{Rueping2009}.

Privacy is a critical consideration while working with individual-level data in different regions and is protected by laws such as the GLBA \cite{GLBA}, HIPAA \cite{HIPAA}, and FERPA \cite{FERPA} in the US and the GDPR \cite{GDPR} in Europe. Failure to appropriately protect patient or customer data exposes organizations to significant reputational and legal risks. Anonymization seeks to protect private and sensitive information of a client and or their activities. When sharing data with privileged partners, stakeholders or regulators, most institutions use popular anonymization techniques (e.g., removal, redaction, encryption, data masking, perturbation, and generalization) to safeguard the privacy and secure the sanctity of the data. 
Application 
Different anonymization techniques distort the dataset in various ways which may be unwanted depending on the downstream task of dataset \cite{Wei}; as anonymization without preserving the data utility, i.e., the information content of the data is not desired.
%
In this paper, we examine the extent to which anonymization techniques preserve subgroups of interest on a dataset. In particular, we investigate the persistence of bias in binary classifiers pre- and post synthesization. Therefore, our utility or figure of merit for an anonymization technique (besides its ability to prevent deanonymization attacks) is its ability to yield the same or similar \textit{most-anomalous} subgroup.


Recent advances in generative machine learning such as variational autoencoders (VAEs) and generative adversarial networks (GANs) are starting to be applied to the anonymization problem \cite{LouizosSLWZ2016, EdwardsS2016, park2018data, bryant2019fairness, Bae, Hakon, Ainsworth}. The main idea of the generative approach is to learn the salient characteristics of the data distribution and sample new (synthetic) individuals from the distribution. Synthetic data from the generative models retains the properties of the original data and can, therefore, be used as a proxy and be shared without risks of re-identification or information leakage \cite{park2018data}.  

We investigate how VAE-based synthesis techniques fare with respect to preserving subgroup properties. We do not pre-specify the subgroups but instead, we use a scanning method that examines various combinations of input attributes to identify the specific subgroup which is consistently misclassified at a rate greater than that of the entire dataset. In particular, we use the \textit{Bias-Scan} method \cite{zhang2017} for anomalous subgroup discovery, a linear-time approach that can be used with large-scale data. We apply Bias-Scan on the original data as well as the synthesized data coming from a VAE and investigate the proportion of the subgroup that was preserved. 
We present our end-to-end pipeline process into a tool what we call Utility Guaranteed Deep Privacy (UGDP).  UGDP can be easily extended to onboard other generative machine learning 
approaches such as GANs to synthesize tabular data. 

 

We use the \emph{Bank of Portugal} dataset \cite{ Moro2014} and the \emph{Adult Dataset} \cite{USAdult} to evaluate our proposed approach. 
Our results indicate two things of note.  First, from the perspective of privacy, the data synthesis procedure does not preserve identities of the most anomalous individual records as shown by the high Jaccard distance index values. We, therefore, conclude 
that the VAE performs quite well in anonymizing the original dataset. Second, from the perspective of utility, the transformation process yields mid-value distance values (about 0.5) for attribute-value overlap indicating that the overall statistical properties of the data are relatively preserved.  

The outline of the remainder of the paper is as follows. In Section~\ref{sec.tools}, we give a brief introduction to VAEs and the Bias-Scan methods as used in the paper. In Section \ref{sec.ugdp}, we present the Utility Guaranteed Deep Privacy (UGDP) system.  Our experiments and empirical results are discussed in Section \ref{sec.results}. Finally, Section \ref{sec.conclusion} concludes the paper and offers some thoughts on future work.  

\section{Preliminaries}
\label{sec.tools}
In this section, we briefly describe the theoretical and algorithmic frameworks that we use for the purposes of this work.

\subsection{Variational Autoencoders}

Autoencoders are generative models designed to capture the underlying distribution of input data and reproducing it using its essential underlying attributes. These essential features are lower-order representations of the data determined within the internal structure of the autoencoder. This lower-order or compressed representation is termed the \textit{latent space}. Note that as information about the original data is compressed, the output of autoencoders is not fundamentally the same as the input ---the output only captures certain aspects of the original data. 
VAEs take this notion of inexact replication of the original data to another level. They are designed to produce variations of the input data, thus creating data that didn't exist, but could have existed based on the underline statistics of that input data. Typically, used for image data reproduction, we are focused on using VAEs to reproduce tabular data as a way to represent the underlying statistics of the input data, thus establishing a data protection mechanism to ensure privacy. 

\subsection{Bias-Scan Algorithm}
This work takes the ``subset scanning'' approach to detecting bias in binary classifiers \cite{zhang2017} and treats the task as a search problem with the goal of finding the subpopulation that is the most systematically over- (or under-) risked by the classifier. This is done by efficiently maximizing a score function $F(S)$ (a likelihood ratio from the Bernoulli distribution) over the exponentially many subgroups.  The efficient maximization is enabled by the Linear Time Subset Scanning property (LTSS) of the scoring function \cite{neill-ltss-2012}.

The categorical features of the dataset create a multi-dimensional tensor with each feature being a mode in the tensor and each record falling into one of the ``cells''.  Bias-Scan identifies an ``axis-aligned'' subset of these cells such that the records in the cells maximize the scoring function.  An example of axis-aligned subsets that span three modes of a tensor would be: Low or Middle income, Black or Hispanic, Females. Bias-Scan iteratively optimizes over each mode of the tensor until convergence to a local maximum is found.  Exploiting the LTSS property of the scoring function guarantees that each optimization step over a mode in the tensor is done efficiently and exactly.  However, the joint optimization over all modes depends on the order in which the modes are optimized, and therefore multiple restarts are used to help explore the space and reach a global maximum.  Consider a feature with 6 possible attribute values.  An exhaustive search over this mode would require $2^{6}$ possible combinations.  However, this maximization can be done by only scoring 6 combinations while still guaranteeing that the optimal subset will be found. 

The scoring function efficiently maximized in Bias-Scan is: 
\[
score_{bias}(S) = max_q  log(q)\sum_{i\in S} y_i - \sum_{i\in S} log(1-\hat{p_i} + q\hat{p_i})
\]
This function is derived from a likelihood ratio of the Bernoulli distribution that operates on a subset of records in the data.  The null hypothesis assumes the binary outcomes $(y_i)$ are drawn correctly from the binary classifier's predicted probability for the records in the subset. The alternative hypothesis assumes that the binary outcomes of the records in the subset are generated by a different probability than predicted by the classifier.  In particular, the alternative hypothesis assumes that the odds $\frac{p}{1-p}$ have been increased by a multiplicative factor $q>1$. The end result of efficiently maximizing this scoring function over all possible ``axis aligned'' subsets of the data is identifying the subgroup that has the most evidence of being generated by the alternative hypothesis stated above.  That is, those records have the largest divergence between what the binary classifier predicted for their outcomes and what was actually observed.


\section{UGDP system design}
\label{sec.ugdp} 

\begin{figure*}[ht]
	\centering
	\includegraphics[width=1\ScaleIfNeeded]{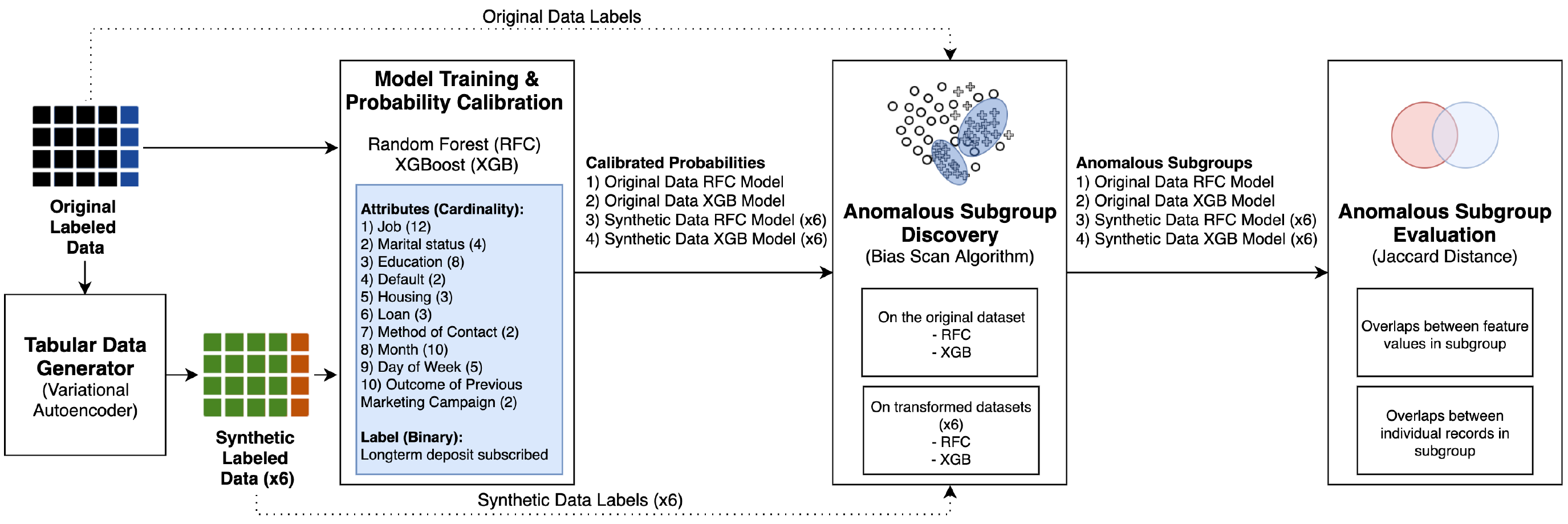}
\caption{Overview of the Utility Guaranteed Deep Privacy (UGDP) System.}
	\label{fig:approach}
\end{figure*}


Our Utility Guaranteed Deep Privacy (UGDP) system identifies the most anomalous subgroup in datasets by maximizing the  bias between the group’s predicted odds ratio from the model and the observed odds ratio from the data and preserve anomalous subgroups when transformed using a variational autoencoder.  Figure \ref{fig:approach}  illustrates an overview of the UGDP system, including: (1) obtaining and pre-processing data, (2) training and calibrating predictive models, (3) employing bias-scan algorithm to discover subgroups on original data, (4) generating tabular data using VAE, (5) employing bias-scan algorithm to discover anomalous subgroup on transformed data, and (6) evaluating the preservation of anomalous subgroup. We now explain each step in more detail.

First, we train two classifiers using the original datasets ---a Random Forest predictive classifier and an XGBoost predictive classifier. Each classifier's parameters were optimized by cross-validation, with model performance measured using Area Under the ROC Curve (AUC). The classifiers were used to generate the predicted (calibrated) probabilities of a respondent taking up the proposed term deposit product. Second, we apply the Bias-Scan algorithm to identify the single subgroup of attribute space with the highest bias score. This is the subgroup of data that is most anomalous with regard to the predicted and the observed outcome. The bias score, therefore, reflects the extent to which the classifier makes the prediction error for the observations in this subgroup.

Third, following \cite{jang2016categorical}, we train a standard categorical variational autoencoder (VAE) to generate new samples of the original dataset. In our case, we use Adaptive Moment Estimation (ADAM) \cite{kingma2014adam} as the optimization method, which computes adaptive learning rates for each parameter with a $LR=0.001$, $\beta _1 = 0.9$, $\beta _2 = 0.999$. Here, the loss function is the reconstruction loss added to the K-L divergence. The input shape of the vectors varies depending on the dataset where all variables were encoded using one-hot encoding. We ran several simulations of the VAE on the original data to obtain 6 different sets of synthetic data samples. 
For each of the 6 synthetic data samples, we trained a Random Forest and an XGBoost model that was then used to obtain the respective predictive probabilities for each new dataset using the binary classifiers. Fourth, we applied the Bias-Scan algorithm to each of the 6 synthetic dataset samples to identify the most anomalous subgroup (subgroup with the highest bias score).

Lastly, for each classifier, we evaluated the pairwise dissimilarity between the individual records belonging to the most anomalous subgroup in the original dataset and the individual records belonging to the most anomalous subgroup in the 6 VAE synthetic samples. We also evaluated the pairwise dissimilarity between the most anomalous subset of attribute values in the original dataset and the most anomalous subset of attribute values in the 6 VAE synthetic samples for each classifier. We quantified pairwise dissimilarity between sets of subgroups using the Jaccard distance, $d_{X,Y}$, defined as a complement of the Jaccard coefficient and obtained by subtracting the Jaccard coefficient from 1: $d_{X,Y} = 1 - \frac{X\cap Y}{X\cup Y}$, where $X$ and $Y$ are sets of discrete elements and  $0$$\leq$$d_{X,Y}$$\leq$$1$ with a higher values implying greater dissimilarity.


\section{Experiment and Empirical Evaluation}
\label{sec.results}

For our experimental evaluation of our approach, we first use the publicly available data from a labeled direct marketing campaign of a Portuguese banking institution consisting of 10 customer attributes (a portion of the original 150) with the binary outcome of the acceptance of an offered long-term bank deposit product. 
This data was collected from May 2008 to November 2010 and consists of 41,188 records with 20 labeled attributes. This data describes whether a customer will accept a long-term bank deposit account (binary target label) \cite{Moro2014}. 
Our preliminary experiment with this dataset takes a subset of the categorical attributes with both nominal and ordinal values. 
Specifically, we performed experimental evaluation on a filtered dataset that included the following 10 discrete attributes (listed with and their cardinality): \textit{Job} (12); \textit{Marital status} (4); \textit{Education} (8); \textit{Default} (2); \textit{Housing} (3); \textit{Loan} (3); \textit{Method of Contact} (2); \textit{Month} (10); \textit{Day of Week} (5); and \textit{Outcome of Previous Marketing Campaign} (2). The final dataset consists of $37,069$ records, 10 attributes, and 1 binary target label.

\begin{table}[ht]
\centering
\begin{tabular}{|c|c|c|}
\cline{1-3}
Model    & Random Forest    & XGBoost \\ \cline{1-3}
Original & 0.85  & 0.83   \\ \cline{1-3}
synthetic\_0 & 0.779 & 0.789  \\ \cline{1-3}
synthetic\_1 & 0.951 & 0.952   \\ \cline{1-3}
synthetic\_2 & 0.946 & 0.949   \\ \cline{1-3}
synthetic\_3 & 0.915 & 0.914   \\ \cline{1-3}
synthetic\_4 & 0.900 & 0.903   \\ \cline{1-3}
synthetic\_5 & 0.970 & 0.969   \\ \cline{1-3}
\end{tabular}
\caption{AUC scores for the Random Forest and XGBoost predictive models on the original Bank of Portugal dataset and six synthetic VAE generated datasets.}
\label{tab:AUCscores}%
\end{table}

\begin{figure}[ht]
	\centering
	\includegraphics[width=1\ScaleIfNeeded]{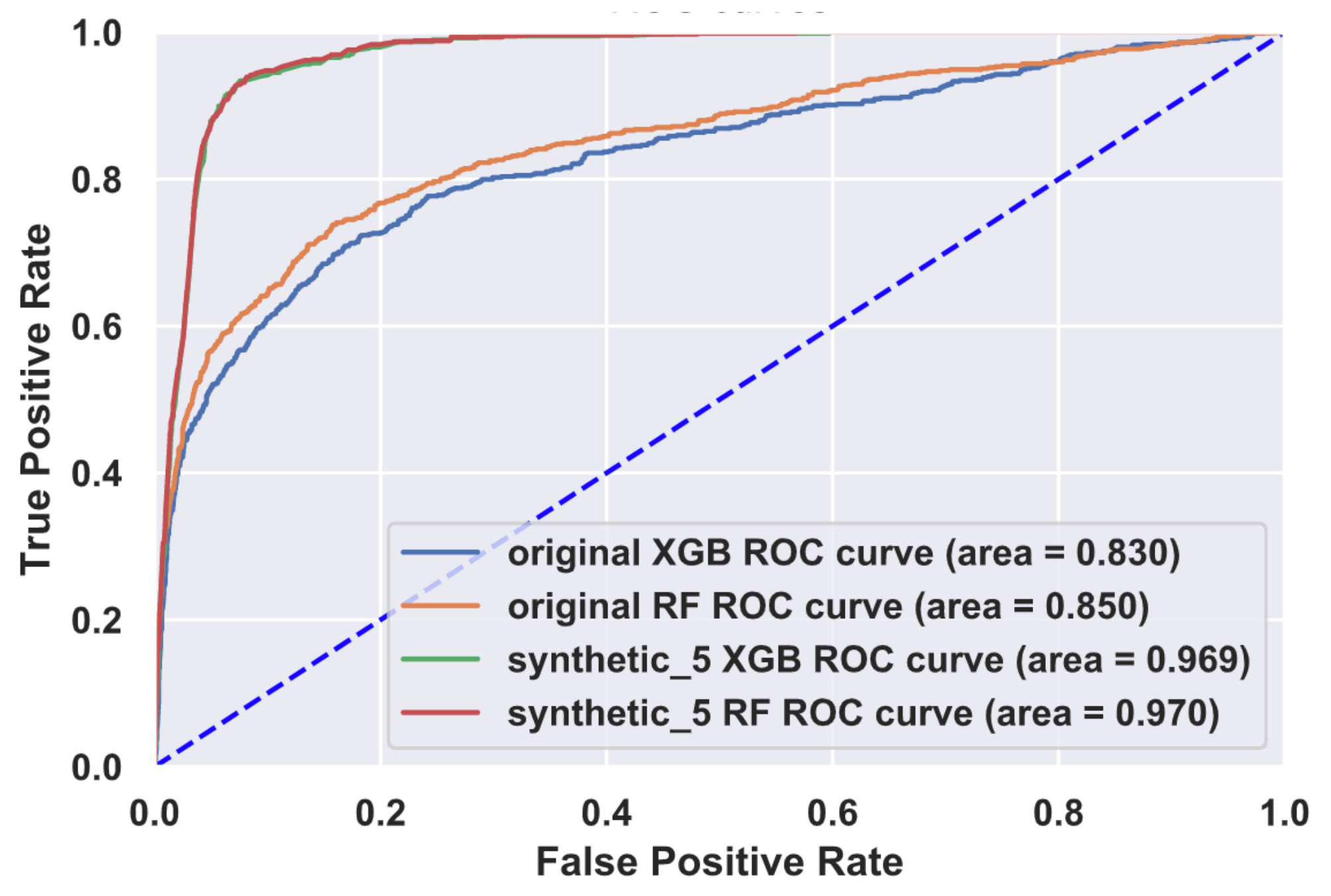}
\caption{Receiver Operating Characteristic (ROC) curves for Random Forest (RF) and XGBoost (XGB) classifiers trained on the original Bank of Portugal dataset and a sample VAE synthetic dataset.}
	\label{fig:ROC_curves}
\end{figure}

\begin{table}[t]
\centering
\begin{tabular}{|c|c|c|c|c|}
\hline
Category & \multicolumn{2}{c|}{Individual Records} & \multicolumn{2}{c|}{Attribute Values} \\ \hline
model    & RF                 & XGB                & RF                & XGB               \\ \hline
synthetic\_0 & 0.965              & 0.953              & 0.571             & 0.556             \\ \hline
synthetic\_1 & 0.949              & 0.958              & 0.481             & 0.536             \\ \hline
synthetic\_2 & 0.959              & 0.958              & 0.516             & 0.543             \\ \hline
synthetic\_3 & 0.949              & 0.975              & 0.469             & 0.545             \\ \hline
synthetic\_4 & 0.947              & 0.951              & 0.438             & 0.441             \\ \hline
synthetic\_5 & 1.000              & 1.000              & 0.462             & 0.581             \\ \hline
\end{tabular}
 \vspace{0.2cm}
    \caption{Dissimilarity, quantified by the Jaccard distance, between subgroups in the original dataset vs 6 VAE synthetic samples that are most affected by bias under different setups. Higher values imply greater dissimilarity}
\label{tab:Jaccard}%
\end{table}
As shown in Table~\ref{tab:AUCscores}, we observe that the two predictive classifiers used in this study have relatively high AUC scores both the original data and the 6 synthetic samples. For example, Figure~\ref{fig:ROC_curves} illustrates the ROC curves for both for the original dataset and a sample VAE generated dataset (synthetic\_5) under both classifiers, and suggests comparable performance of the random forest and XGBoost models for each dataset.

The results of our investigation of the overlaps between the anomalous subgroups of individual records and attribute values in the original data and the VAE synthetic datasets are shown in Table~\ref{tab:Jaccard}. These results show the impact of the VAE transformation on the subgroup predictive bias of the classifiers. The Jaccard distances for individual records indicate little to no overlap between individual records of the subgroup most affected by the predictive bias of Random Forest and XGBoost classifiers. The Jaccard distances for the attribute values suggest considerable overlap between subsets of the attribute space that are most affected by predictive bias. 

\begin{figure}[ht]
\begin{minipage}[b]{1.0\linewidth}
  \centering
  \centerline{	\includegraphics[width=0.6\ScaleIfNeeded]{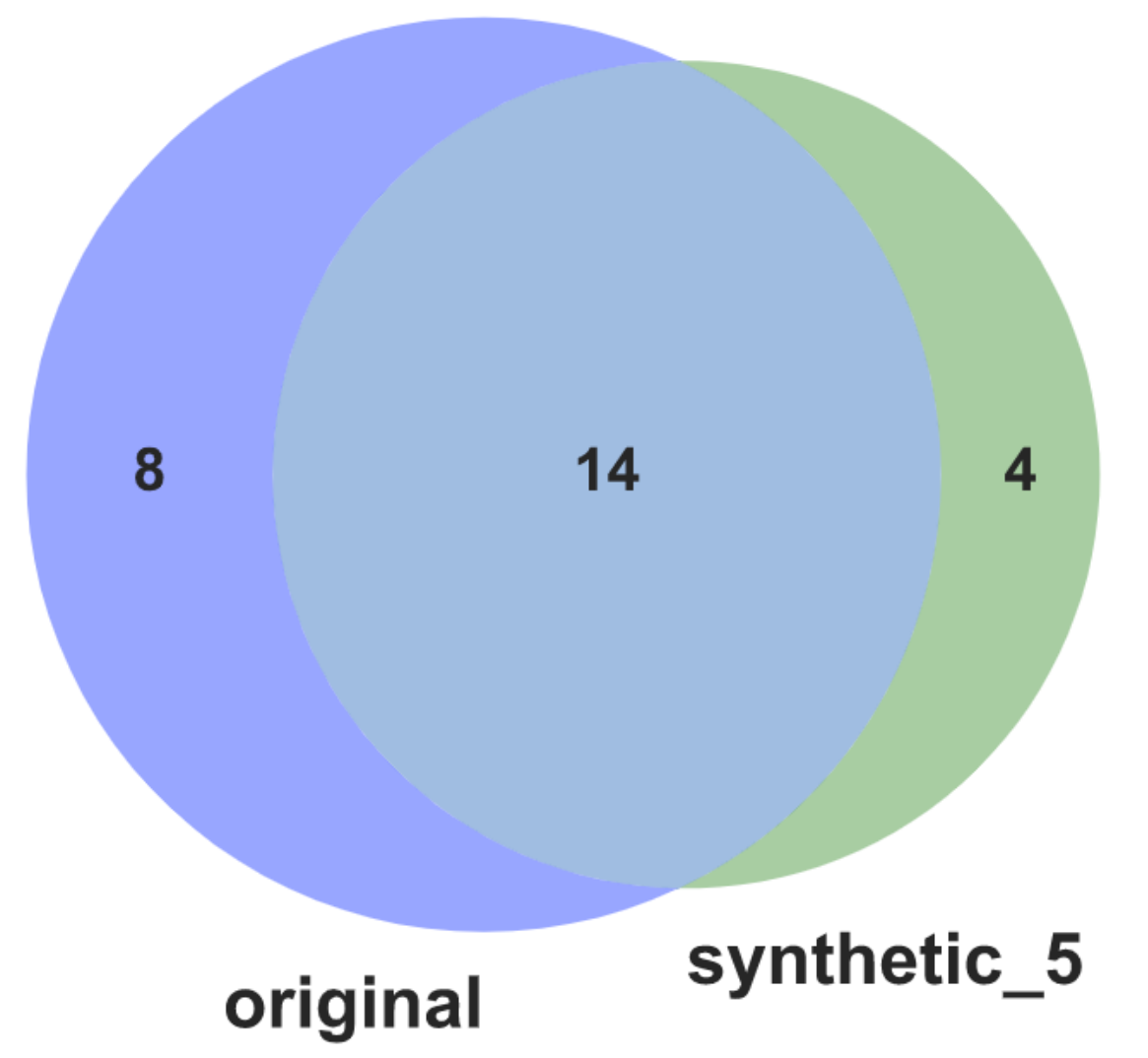}}
  \centerline{(a)}\medskip
\end{minipage}

\hfill
\begin{minipage}[b]{1.0\linewidth}
  \centering
  \centerline{\includegraphics[width=0.6\ScaleIfNeeded]{original_synthetic_5_Feature_Overlap_RFC_Model_v2.pdf}}
  \centerline{(b)}\medskip
\end{minipage}
\caption{(a) shows that for a Random Forest classifier trained on the original Bank of Portugal data and on a synthetic data generated by the VAE, with no overlap between individual records of the subgroup most affected by the predictive bias of the classifier. (b) shows that for the same setup, there is considerable overlap (Jaccard distance = 0.462).
}
\label{fig:overlaps}
\end{figure}

This observation is exemplified in Figure~\ref{fig:overlaps} which compares the overlap between the most anomalous subgroups in the original dataset and a sample VAE generated synthetic dataset (synthetic\_5) under the random forest classifier. As shown in Figure~\ref{fig:overlaps}(a), there is no overlap (Jaccard distance = 1) between the individual records of the subgroup most affected by the predictive bias of the classifier as measured by the Jaccard distance metric which shows values very close to one. This suggests that the VAE performs relatively well in data anonymization and re-identification of individuals is not guaranteed. 
However, we observe that for the attribute (feature) values, there is a significant overlap (Jaccard distance = 0.462) between the most anomalous subgroups of the model as shown in Figure~\ref{fig:overlaps}(b). This implies that the characteristics of subgroups of interest in a dataset would not be lost when the data is transformed using VAEs.

Finally, we further tested the UGDP system using our second dataset, namely the Adult Dataset \cite{USAdult} (the 1994 (United States) Census-bureau dataset), to validate the approach with respect to preserving subgroup properties and privacy on transformed data through synthesis techniques.
This dataset consists of 7 attributes from the original 14 features that are used to predict whether a person's income is higher or lower than \$50k per year threshold. In particular, we observed minimal overlap between the individual records of the subgroups most affected by the predictive bias of the Random Forest (Jaccard distance = 0.96) and the XGB model (Jaccard distance = 0.89). Similarly to the results observed in the Bank of Portugal data experiment, there was more overlap in the attribute values of the most anomalous subgroup for the Random Forest model (Jaccard distance = 0.65) and the XGB model (Jaccard distance = 0.43).

\section{conclusion}\label{sec.conclusion}

In this paper, we have presented the standard VAE as a tool for anonymization and utility preservation. In particular, we have demonstrated that this model achieves anonymization on individual records which allows for data reuse and sharing amongst the key stakeholders and clients. This is confirmed by the mid-value Jaccard distance on the overlaps for the anomalous subsets of individual records for the original and synthetic datasets after the data has been run through a classifier and the Bias-Scan algorithm. We have also shown that the VAE synthesizer preserves \textit{some}`global' statistical distributional properties of the data as demonstrated by the mid-level values of the Jaccard distance metric for the feature values. We therefore conclude that the data transformation and synthesization preserves (to an extent) some key subgroup properties of the data.   

A potential direction for further extensions of this work could be on developing a disciplined approach that can attain higher subgroup overlaps of attribute values between the two data sets while still achieving the high privacy levels as represented by minimal overlap at the level of the individual records. An initial approach would be modifying the standard VAE for data generation to a case where we incorporate constraints on how each attribute can vary with respect to one another when generating the synthetic data from the VAE. We are also exploring alternative generative approaches such as GANs~\cite{park2018data, kumar2018ecommercegan} to synthesize tabular data and check for consistency values between different attributes while preserving the privacy of individual records.

\bibliographystyle{IEEEbib}
\bibliography{icassp}

\end{document}